\documentclass{article}

\usepackage{microtype}
\usepackage{graphicx}
\usepackage{booktabs} 

\usepackage{amsmath}
\usepackage{amsfonts}
\usepackage{multirow}
\usepackage{makecell}
\usepackage{rotating}
\newcommand{\STAB}[1]{\begin{tabular}{@{}c@{}}#1\end{tabular}}
\usepackage{subcaption}
\usepackage[font=bf,labelfont=bf]{caption}

\usepackage{hyperref}

\newcommand{\x}{\mathbf{x}}
\newcommand{\X}{\mathbf{X}}

\DeclareMathOperator{\PMI}{PMI}
\DeclareMathOperator{\Diff}{Diff}

\newcommand{\ie}{\textit{i}.\textit{e}., }

\usepackage[accepted]{icml2021}

\icmltitlerunning{Informative Class Activation Maps}

\begin{document}

\twocolumn[
\icmltitle{
Informative Class Activation Maps
}

\icmlsetsymbol{equal}{*}

\begin{icmlauthorlist}
\icmlauthor{Zhenyue Qin}{equal,anu}
\icmlauthor{Dongwoo Kim}{equal,anu,postech}
\icmlauthor{Tom Gedeon}{anu}
\end{icmlauthorlist}

\icmlaffiliation{anu}{School of Computing, Australian National University}
\icmlaffiliation{postech}{GSAI, PosTech, Korea}

\icmlcorrespondingauthor{Zhenyue Qin}{zhenyue.qin@anu.edu.au}
\icmlcorrespondingauthor{Dongwoo Kim}{dongwoo.kim@anu.edu.au}


\vskip 0.3in
]




\printAffiliationsAndNotice{\icmlEqualContribution} 

\begin{abstract}
We study how to evaluate the quantitative information content of a region within an image for a particular label. To this end, we bridge class activation maps with information theory. We develop an informative class activation map (infoCAM). Given a classification task, infoCAM depict how to accumulate information of partial regions to that of the entire image toward a label. Thus, we can utilise infoCAM to locate the most informative features for a label. 
When applied to an image classification task, infoCAM performs better than the traditional classification map in the weakly supervised object localisation task. We achieve state-of-the-art results on Tiny-ImageNet. 
\end{abstract}

\section{Introduction}

Class activation maps are useful tools for identifying regions of an image corresponding to particular labels. Many weakly supervised object localization methods are based on class activation maps (CAMs). In this paper, we formalize CAMs to be related to information theory. Furthermore, with the new view of neural network classifiers as mutual information evaluators, we are able to depict the quantitative relationship between the information of the entire image and its local regions about a label. We call our new CAM Informative Class Activation Map (infoCAM), since it is based on information theory. Moreover, infoCAM can also improve the performance of the weakly supervised object localisation (WSOL) task than the original CAM.

\section{Informative Class Activation Map}

To explain infoCAM, we first introduce the concept and definition of the class activation map. We then show how to apply it to weakly supervised object localisation (WSOL). 

\begin{figure*}[t!]
    \centering
    \includegraphics[width=0.83\linewidth]{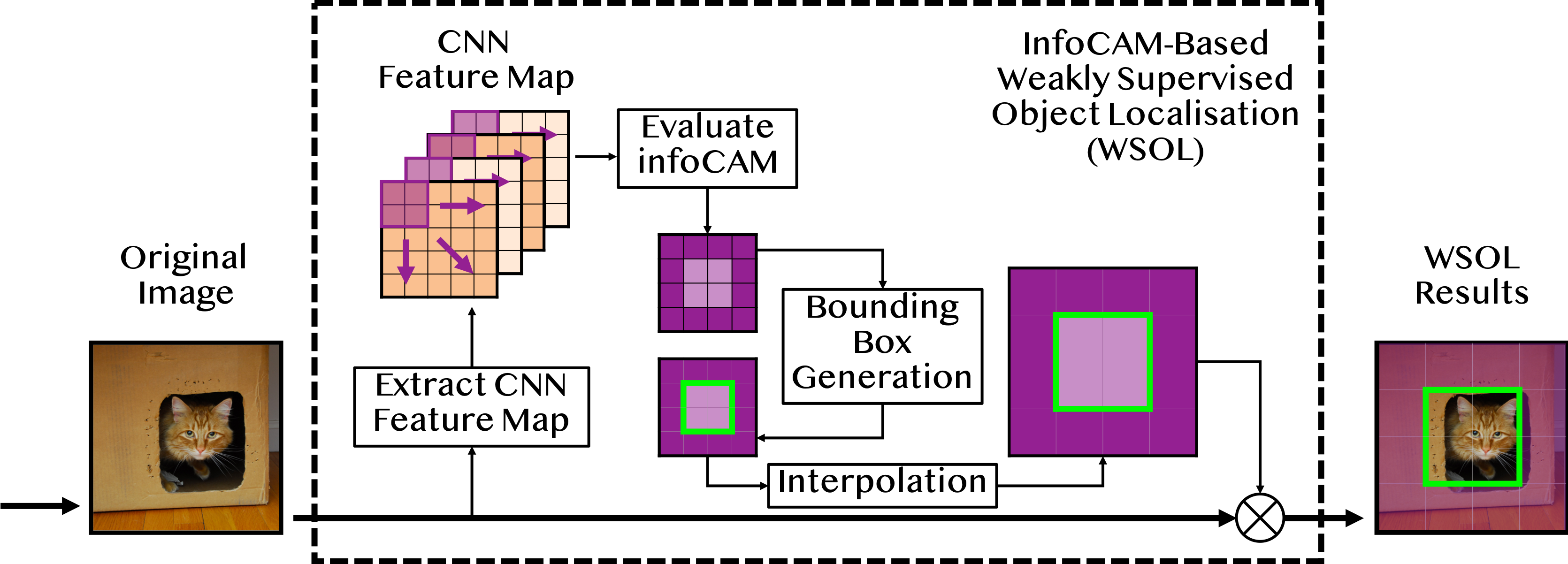}
    \caption{A visualization of the infoCAM procedure for the WSOL task. The task aims to draw a bounding box for the target object in the original image. The procedure includes: 1) feed input image into a CNN to extract its feature maps, 2) evaluate PMI difference between the true and the other labels of input image for each region within the feature maps, 3) generate the bounding box by keeping the regions exceeding certain infoCAM values and find the largest connected region and 4) interpolate and map the bounding box to the image.}
    \label{fig:infoCAM-Illustration}
\end{figure*}

\subsection{CAM: Class Activation Map}

Contemporary classification convolutional neural networks (CNNs) consist of stacks of convolutional layers interleaved with pooling layers for extracting visual features. 
These convolutional layers result in feature maps. A feature map is a collection of 2-dimensional grids. The size of the feature map depends on the structure of convolution and pooling layers. Generally the feature map is smaller than the original image. The number of feature maps corresponds to the number of convolutional filters.
The feature maps from the final convolutional layer are usually averaged, flattened and fed into the fully-connected layer for classification~\cite{lin2014network}.
Given $K$ feature maps $g_1, .. , g_K$, the fully-connected layer consists of weight matrix $W \in \mathbb{R}^{M \times K}$, where $w_k^y$ represents the scalar weight corresponding to class $y$ for feature $k$.
We use $g_k(a,b)$ to denote a value of 2-dimensional spatial point $(a,b)$ with feature $k$ in map $g_k$.
In~\cite{choe2019attention}, the authors propose a way to interpret the importance of each point in feature maps. The importance of spatial point $(a,b)$ for class $y$ is defined as a weighted sum over features:
\begin{align}
\label{eq:cam_def}
    M_{y}(a, b) = \sum_{k} w_k^{y} g_{k} (a, b).
    \vspace{-4mm}
\end{align}
 We redefine $M_{y}(a, b)$ as an intensity of the point $(a, b)$.
The collection of these intensity values over all grid points forms a class activation map (CAM). CAM highlights the most relevant region in the feature space for classifying $y$.
The input going to the softmax layer corresponding to the class label $y$ is: 
\begin{align}
    \sum_{a,b} M_{y}(a, b) = n(\x)_y.
    \vspace{-4mm}
\end{align}
Intuitively, weight $w_k^y$ indicates the overall importance of the $k$th feature  to class $y$, and intensity $M_{y}(a, b)$ implies the importance of the feature map at spatial location $(a, b)$ leading to the classification of image $\mathbf{x}$ to $y$.

\textbf{WSOL}: The aim of WSOL is to identify the region containing the target object in an image given a label, without any pixel-level supervision.
Previous approaches tackle WSOL by creating a bounding box from the CAM~\cite{choe2019attention}. Such a CAM contains all important locations that exceed a certain intensity threshold. The box is then upsampled to match the size of the original image.

\subsection{InfoCAM: Informative Class Activation Map}
In~\cite{qin2019rethinking}, the authors show that softmax classifier carries an explicit implication between inputs and labels in terms of information theory. We extend the notion of mutual information from being a pair of an input image and a label to regions of the input image and labels to capture the regions that have high mutual information with labels.

To simplify the discussion, we assume here that there is only one feature map, \ie $K=1$. However, the following results can be easily applied to the general cases where $K>1$ without loss of generality.
We introduce a region $R$ containing a subset of grid points in feature map $g$. 

Mutual information is an expectation of the point-wise mutual information (PMI) between two variables, \ie $\mathbb{I}(\X,Y) = \mathbb{E}_{\x,y}[\text{PMI}(\x,y)]$. Given two instances of variables, we can estimate their PMI of $\x$ and $y$ as:
\begin{align}
\PMI(\x, y) = n(\x)_y - \log\sum_{y'=1}^{M}\exp(n(\x)_{y'}) + \log M. \notag
\end{align}
The PMI is close to $\log M$ if $y$ is the maximum argument in log-sum-exp. To find a region which is the most beneficial to the classification, we compute the difference between PMI with true label and the average of the other labels and decompose it into a point-wise summation as
\begin{align}
\Diff(\PMI(\x)) = \PMI(\x, y^*) - \frac{1}{M-1}\sum_{y' \neq y^*}\PMI(\x, y') \notag\\
= \sum_{(a,b)\in g} w^{y*} g(a,b) - \frac{1}{M-1}\sum_{y' \neq y^*} w^{y'} g(a,b). \notag
\end{align}
The point-wise decomposition suggests that we can compute the PMI differences with respect to a certain region. Based on this observation, we propose a new CAM, named informative CAM or infoCAM, with the new intensity function $M_{y}^{\Diff}(R)$ between region $R$ and label $y$ defined as follows:
\begin{align*}
M_{y}^{\Diff}(R) = \sum_{(a,b)\in R} w^yg(a,b) - \frac{1}{M-1}\sum_{y' \neq y}w^{y'}g(a,b).
\label{eq:info_cam_correct}
\vspace{-2mm}
\end{align*}
The infoCAM highlights the region which decides the classification boundary against the other labels. 
Moreover, we further simplify the above equation
to be the difference between PMI with the true and the most-unlikely labels according to the classifier's outputs, denoting as infoCAM+, with the new intensity: 
\begin{align}
M_{y}^{\Diff^+}(R) = \sum_{(a,b)\in R} w^y g(a,b) - w^{y'}g(a,b), 
\end{align}
where $y' = \underset{m}{\arg \min} \sum_{(a,b)\in R} w^{m}g(a,b)$. 

The complete procedure of WSOL with infoCAM is visually illustrated in \autoref{fig:infoCAM-Illustration}. 
\section{Object Localisation with InfoCAM}
\label{sec:exp}
In this section, we demonstrate experimental results with infoCAM for WSOL. We first describe the experimental settings and then present the results.

\begin{table}[t!]
\centering
\resizebox{0.45\textwidth}{!}{
\begin{tabular}{c l r r r r}
\toprule
&  &  \multicolumn{2}{c}{CUB-200-2011} & \multicolumn{2}{c}{Tiny-ImageNet}\\
&  & 
\makecell{GT \\ Loc. (\%)} & \makecell{Top-1 \\ Loc. (\%)} & \makecell{GT \\ Loc. (\%)} & \makecell{Top-1 \\ Loc. (\%)}\\ 
\midrule
\multirow{6}{*}{\STAB{\rotatebox[origin=c]{90}{VGG}}} 
& CAM& 42.49 & 31.38 & 53.49 & 33.48 \\
& CAM (ADL) & 71.59 & 53.01 &  52.75 & 32.26 \\
& {infoCAM} & {52.96} & {39.79} & {55.50} & {34.27}\\
& {infoCAM (ADL)} & {73.35} & {53.80} & {53.95} & {33.05} \\ 
& {infoCAM+} & {59.43} & {44.40} &  {55.25} & {34.27}\\
& {infoCAM+ (ADL)} & \textbf{75.89} & {54.35} &  {53.91} & {32.94} \\  \midrule
\multirow{6}{*}{\STAB{\rotatebox[origin=c]{90}{ResNet}}} 
& CAM& 61.66 & 50.84 &  54.56 & 40.55  \\
& CAM (ADL) & 57.83 & 46.56 &  52.66 & 36.88  \\
& {infoCAM} &  {64.78} & {53.22} & \textbf{57.79} & \textbf{43.34} \\
& {infoCAM (ADL)} & {67.75} & {54.71} & {54.18} & {37.79} \\
& {infoCAM+} &  {68.99} & \textbf{55.83} &  {57.71} & {43.07} \\
& {infoCAM+ (ADL)} & {69.63} & {55.20} &  {53.70} & {37.71} \\
\bottomrule
\end{tabular}
}
\vspace{1em}
\caption{Localisation results of CAM and infoCAM on CUB-2011-200 and Tiny-ImageNet. InfoCAM outperforms CAM on localisation of objects with the same model architecture. Bold values represent the highest accuracy for a certain metric. }
\label{table:cam_info_cam}
\end{table}

\subsection{Experimental settings}
We evaluate WSOL performance on CUB-200-2011~\cite{wah2011caltech} and Tiny-ImageNet~\cite{tiny_imagenet}. CUB-200-2011 consists of 200 bird specifies. 
Since the dataset only depicts birds, not including other kinds of objects, variations due to class difference are subtle~\cite{dubey2018pairwise}. 
Such nuance-only detection can lead to localisation accuracy degradation~\cite{choe2019attention}. 

Tiny-ImageNet is a reduced version of ImageNet.
Compared with the full ImageNet, training classifiers on Tiny-ImageNet is faster due to image resolution reduction and quantity shrinkage, yet classification becomes more challenging~\cite{odena2017conditional}.

\begin{figure}[t!]
    \centering
    \includegraphics[width=0.76\linewidth]{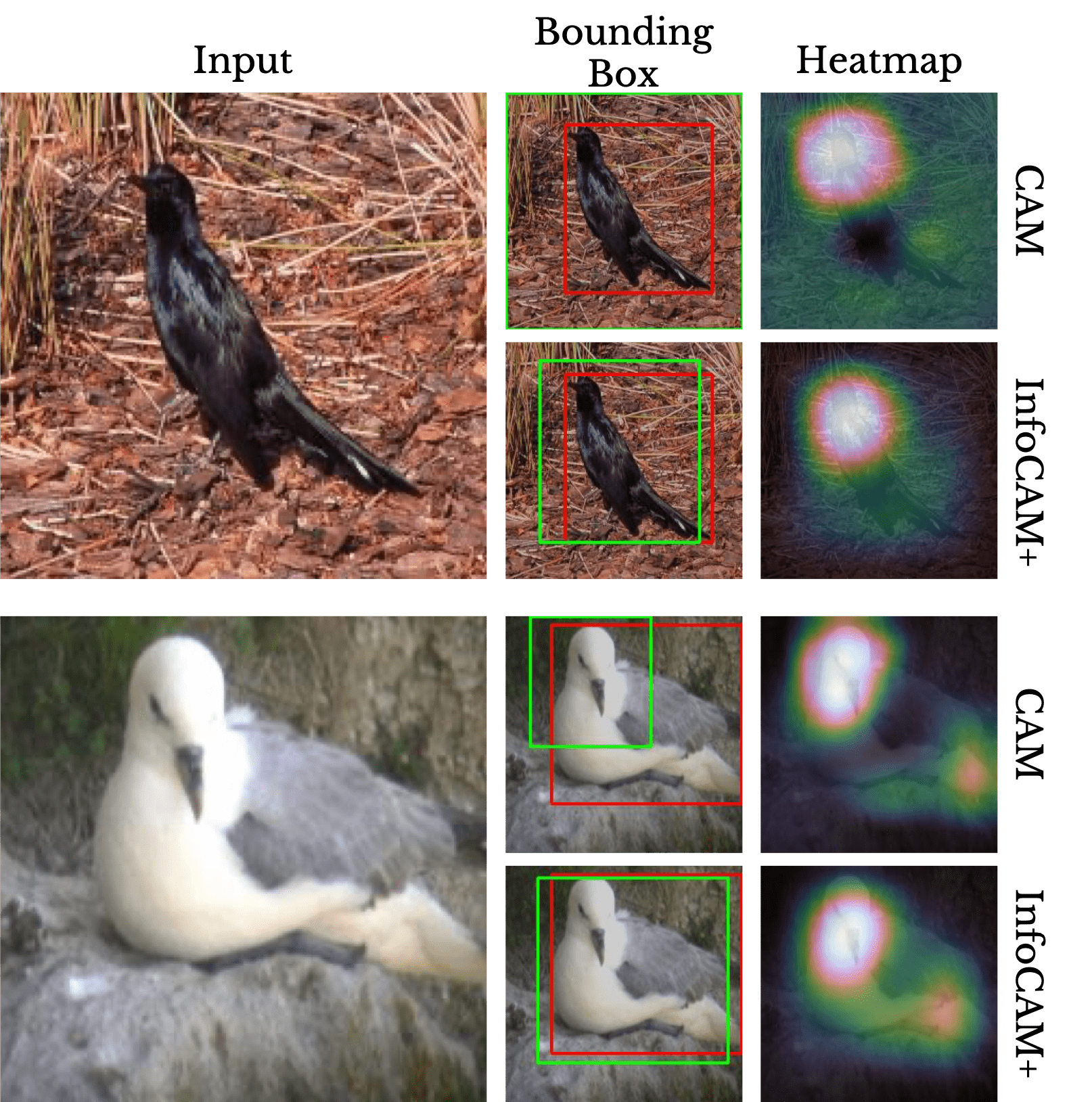}
    \caption{Visualisation of comparison between CAM and infoCAM+. Red and green boxes represent the ground truth and prediction, respectively. Brighter regions represent higher CAM or infoCAM+ values.}
    \label{fig:reg-sig}
\end{figure}

To perform an evaluation on localisation, we first need to generate a bounding box for the object within an image. We generate a bounding box in the same way as in~\cite{zhou2016learning}. That is, after evaluating infoCAM within each region of an image, we only retain the regions whose infoCAM values are more than 20\% of the maximum infoCAM and abandon all the other regions. Then, we draw the smallest bounding box that covers the largest connected component. 

We follow the same evaluation metrics in~\cite{choe2019attention} to evaluate localisation performance with two accuracy measures: 1) localisation accuracy with known ground truth class (GT Loc.), and 2) top-1 localisation accuracy (Top-1 Loc.). GT Loc. draws the bounding box from the ground truth of image labels, whereas Top-1 Loc. draws the bounding box from the predicted most likely image label and also requires correct classification. The localisation of an image is judged to be correct when the intersection over union of the estimated bounding box and the ground-truth bounding box is greater than 50\%.

\begin{figure*}[t!]
    \centering
    \includegraphics[width=0.82\linewidth]{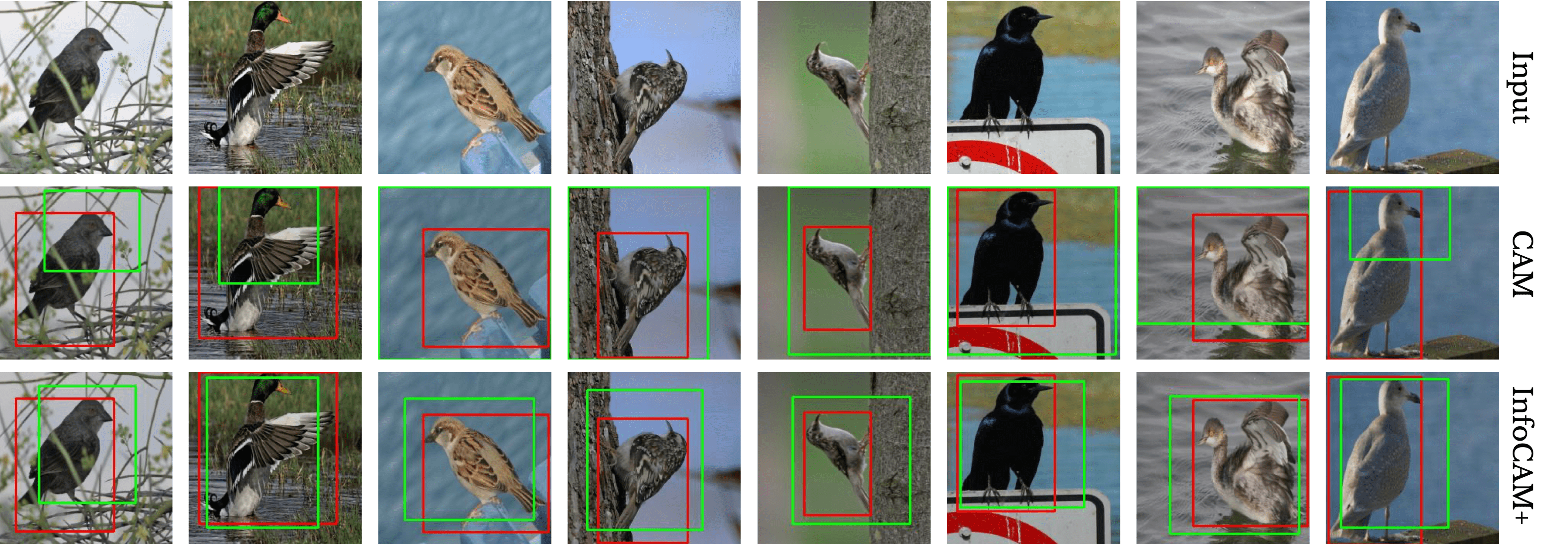}
    \caption{Visualisation of localisation with ResNet50 without using ADL on CUB-200-2011. Images in the second and the third row correspond to CAM and infoCAM+, respectively. Estimated (green) and ground-truth (red) bounding boxes are shown separately.}
    \label{fig:cam-illustration-butterfly}
\end{figure*}

We adopt the same network architectures and hyper-parameters as in~\cite{choe2019attention}, which shows the current state-of-the-art performance. Specifically, the network backbone is ResNet50~\cite{he2016deep} and a variation of VGG16~\cite{szegedy2015going}, in which the fully connected layers are replaced with global average pooling (GAP) layers to reduce the number of parameters. The traditional softmax is used as the final layer since both datasets are well balanced.
InfoCAM requires the region parameter $R$. We apply a square region for the region parameter $R$. The size of the region $R$ is set as $5$ and $4$ for VGG and ResNet in CUB-200-2011, respectively, and $3$ for both VGG and ResNet in Tiny-ImageNet.

These models are tested with the Attention-based Dropout Layer (ADL) to tackle the localisation degradation problem~\cite{choe2019attention}. ADL is designed to randomly abandon some of the most discriminative image regions during training to ensure CNN-based classifiers cover the entire object. The ADL-based approaches demonstrate state-of-the-art performance in CUB-200-2011~\cite{choe2019attention} and Tiny-ImageNet~\cite{choe2018improved} for the WSOL task and are computationally efficient. We test ADL with infoCAMs to enhance WSOL capability.

To prevent overfitting in the test dataset, we evenly split the original validation images to two data piles, one still used for validation during training and the other acting as the final test dataset. We pick the trained model from the epoch that demonstrates the highest top-1 classification accuracy in the validation dataset and report the experimental results with the test dataset. All experiments are run on two Nvidia 2080-Ti GPUs, with the PyTorch deep learning framework~\cite{paszke2017automatic}. 

\subsection{Experimental Results}

Table~\ref{table:cam_info_cam} shows the localisation results on CUB-200-2011 and Tiny-ImageNet. The results demonstrate that infoCAM can consistently improve accuracy over the original CAM for WSOL under a wide range of networks and datasets. Both infoCAM and infoCAM+ perform comparably to each other. ADL improves the performance of both models with CUB-200-2011 datasets, but it reduces the performance with Tiny-ImageNet. We conjecture that dropping any part of a Tiny-ImageNet image with ADL significantly influences classification since the images are relatively small.

\autoref{fig:reg-sig} highlights the difference between CAM and infoCAM. The figure suggests that infoCAM gives relatively high intensity on the object to compare with that of CAM, which only focuses on the head part of the bird.
Figure~\ref{fig:cam-illustration-butterfly} in the Appendix presents additional examples of visualisation for comparing localisation performance of CAM to infoCAM, both without the assistance of ADL
. From these visualisations, we notice that 
the bounding boxes generated from infoCAM are formed closer to the objects than the original CAM. That is, infoCAM tends to precisely cover the areas where objects exist, with almost no extraneous or lacking areas. For example, CAM highlights the bird heads, whereas infoCAM also covers the bird bodies. 


\subsection{Localisation of multiple objects with InfoCAM}
\begin{figure}[ht]
\centering
\begin{subfigure}{.14\textwidth}
  \centering
  \includegraphics[width=.99\linewidth]{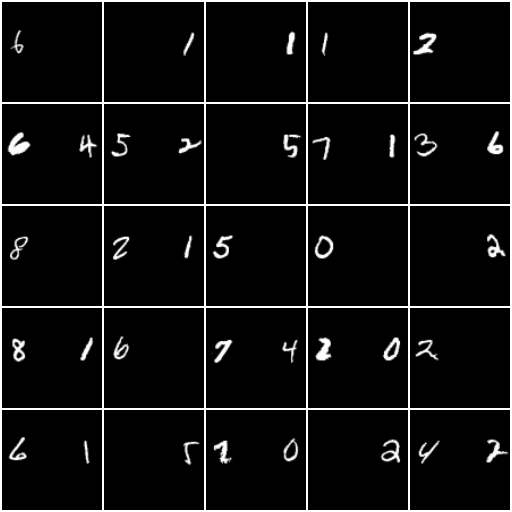}  
  \caption{Original}
  \label{fig:sub-first}
\end{subfigure}
\begin{subfigure}{.14\textwidth}
  \centering
  \includegraphics[width=.99\linewidth]{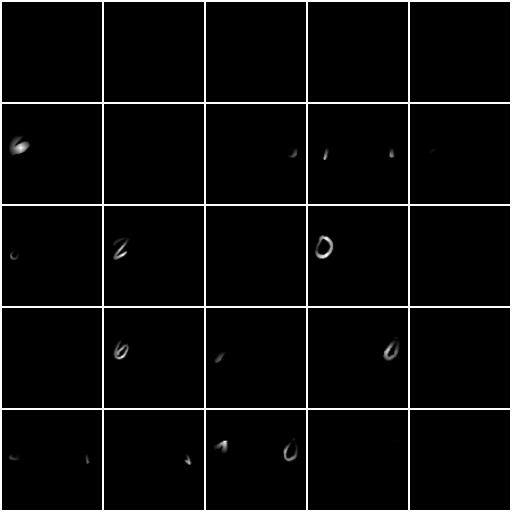}  
  \caption{CAM}
  \label{fig:sub-second}
\end{subfigure}
\begin{subfigure}{.14\textwidth}
  \centering
  \includegraphics[width=.99\linewidth]{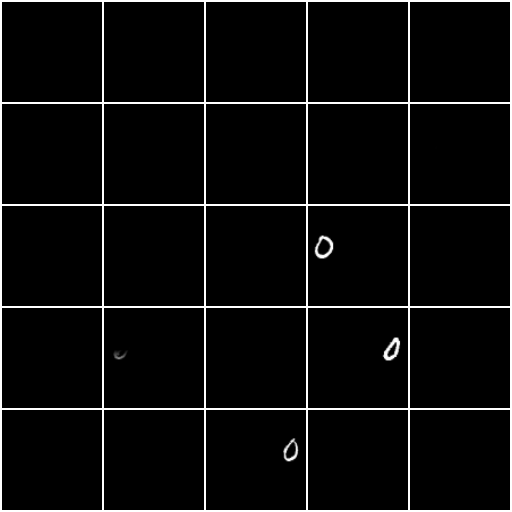}  
  \caption{InfoCAM }
  \label{fig:sub-second}
\end{subfigure}
\caption{Visualisation of comparison between CAM and infoCAM for the multi-MNIST dataset. Each image has one or two digits in the left and/or right. We aim to extract digit 0. }
\label{fig:multi_mnist}
\end{figure}
So far, we have shown the results of localisation from a multi-class classification problem. We further extend our experiments on localisation to multi-label classification problems. A softmax function is a generalisation of its binary case, a sigmoid function. Therefore, we can apply infoCAM to each label for a multi-label classification problem, which is a collection of binary classification tasks.

For the experiment, we construct a double-digit MNIST dataset where each image contains up to two digits randomly sampled from the original MNIST dataset~\cite{lecun2010mnist}. We locate one digit on the left-side, and the other on the right-side. Some of the images only contain a single digit. For each side, we first decide whether to include a digit from a Bernoulli distribution with mean of 0.7. Then each digit is randomly sampled from a uniform distribution. However, we remove the images that contain no digits. 
Random samples from the double-digit MNIST are shown in \autoref{fig:sub-first}.

\begin{table}[t!]
\centering
\resizebox{0.48\textwidth}{!}{
\begin{tabular}{c l l l l l l l l l l}
\toprule
\multirow{2}{*}{\makecell{Type}} & \multicolumn{10}{c}{Digit Classification Accuracy (\%)} \\
& 0 & 1 & 2 & 3 & 4 & 5 & 6 & 7 & 8 & 9 \\
\midrule
sigmoid & 1.00 & 0.84 & 0.86 & 0.94 & 0.89 & 0.87 & 0.87 & 0.86 & 1.00 & 1.00 \\
PC-sigmoid & 1.00 & 1.00 & 1.00 & 1.00 & 1.00 & 1.00 & 1.00 & 1.00 & 1.00 & 1.00 \\
\bottomrule
\end{tabular}
}
\caption{Comparison between the classification accuracy results with sigmoid and PC-sigmoid on the double-digit MNIST dataset. }
\vspace{-4mm}
\label{tbl:sig_pc_sig_acc}
\end{table}

We first compare the classification accuracy results between using the original sigmoid and PC-sigmoid. 
As shown in \autoref{tbl:sig_pc_sig_acc}, PC-sigmoid increases the classification accuracy for each digit type on the test set.
InfoCAM improves the localisation accuracy for the WSOL task as well. CAM achieves the localisation accuracy of 91\%. InfoCAM enhances the localisation accuracy to 98\%. Qualitative visualizations are displayed in \autoref{fig:multi_mnist}. We aim to preserve the regions of an image that are most relevant to a digit, and erase all the other regions. From the visualization, one can see that infoCAM localizes digits more accurately than CAM.

\section{Conclusion}
In~\cite{qin2019rethinking}, the authors convert neural network classifiers to mutual information estimators. Then, using the pointwise mutual information between the inputs and labels, we can locate the objects within images more precisely. We also provide a more information-theoretic interpretation of class activation maps. Experimental results demonstrate the effectiveness of our proposed method.

\newpage
\nocite{langley00}

\bibliography{refs}
\bibliographystyle{icml2021}

\appendix
\newpage
\onecolumn
\section{Further Result}
In this section, we present some further results on localisation and classification.

\subsection{Localisation and Classification Result}

\autoref{table:cam_info_cam_full} is a reproduction of main result with the classification results. 
Note that the classification performances of CAM and infoCAM is the same since we do not modify the training objective of infoCAM. The result can be used to understand the effect of ADL on the classification task.

\begin{table*}
\centering
\begin{tabular}{c l c c c c}
\toprule
&  & 
\makecell{GT \\ Loc. (\%)} &
\makecell{Top-1 \\ Loc. (\%)} & 
\makecell{Top-1 \\ Cls (\%)} & 
\makecell{Top-5 \\ Cls (\%)}  \\ 
\midrule
\multirow{6}{*}{\makecell{VGG-\\16-\\GAP}} & CAM& 42.49 & 31.38 & 73.97 & 91.83   \\
& CAM (ADL) & 71.59 & 53.01 & 71.05 & 90.20 \\
& {infoCAM} & {52.96} & {39.79} & - & - \\
& {infoCAM} (ADL) & {73.35} & {53.80} & - & - \\
& {infoCAM+} & {59.43} & {44.40} & - & - \\
& {infoCAM+} (ADL) & \textbf{75.89} & {54.35} & - & - \\ \midrule
\multirow{6}{*}{\makecell{ResNet-\\50}} & CAM& 61.66 & 50.84 & 80.54 & 94.09 \\
& CAM (ADL) & 57.83 & 46.56 & 79.22 & 94.02 \\
& {infoCAM} &  {64.78} & {53.22} & - & -  \\
& {infoCAM} (ADL) & {67.75} & {54.71} & - & -  \\
& {infoCAM+} &  {68.99} & \textbf{55.83} & - & -  \\
& {infoCAM+} (ADL) & {69.63} & {55.20} & - & -  \\
\bottomrule
\end{tabular}

\caption{Evaluation results of CAM and infoCAM on CUB-2011-200. Note that the classification accuracy of infoCAM is the same as those of CAM. InfoCAM always outperforms CAM on localisation of objects under the same model architecture.}
\label{table:cam_info_cam_full}
\end{table*}

\begin{table*}
\centering

\centering
\begin{tabular}{c l c c c c}
\toprule
 &  & 
\makecell{GT \\ Loc. (\%)} &
\makecell{Top-1 \\ Loc. (\%)} & 
\makecell{Top-1 \\ Cls (\%)} & 
\makecell{Top-5 \\ Cls (\%)}  \\ 
\midrule
\multirow{6}{*}{\makecell{VGG-\\16-\\GAP}} & CAM& 53.49 & 33.48 & 55.25 & 79.19  \\
& CAM (ADL) & 52.75 & 32.26 & 52.48 & 78.75 \\
& {infoCAM} & {55.50} & {34.27} & - & -\\
& {infoCAM} (ADL) & {53.95} & {33.05} & - & - \\ 
& {infoCAM+} & {55.25} & {34.27} & - & -\\
& {infoCAM+} (ADL) & {53.91} & {32.94} & - & - \\ \midrule
\multirow{6}{*}{\makecell{ResNet-\\50}} & CAM& 54.56 & 40.55 & 66.45 & 86.22 \\
& CAM (ADL) & 52.66 & 36.88 & 63.21 & 83.47 \\
& {infoCAM}& \textbf{57.79} & \textbf{43.34} & - & - \\
& {infoCAM} (ADL) & {54.18} & {37.79} & - & - \\
& {infoCAM+}& {57.71} & {43.07} & - & - \\
& {infoCAM+} (ADL) & {53.70} & {37.71} & - & - \\
\bottomrule
\end{tabular}

\caption{Evaluation results of CAM and infoCAM on Tiny-ImageNet. Note that the classification accuracy of infoCAM is the same as those of CAM. InfoCAM always outperforms CAM on localisation of objects under the same model architecture.}
\label{table:cam_info_cam_full}
\end{table*}

\subsection{Ablation Study}

\autoref{tbl:ablation} shows the result of ablation study. We have tested the importance of three features: 1) ADL, 2) region parameter $R$ and 3) the second subtraction term in the infoCAM equation. To combine the result in the main text, the result suggests that both region parameter and subtraction term are necessary to increase the performance of localisation. The choice of ADL depends on the dataset. We conjecture that ADL is inappropriate to apply Tiny-ImageNet since the removal of any part of tiny image, which is what ADL does during training, affects the performance of the localisation to compare with its application to relatively large images.

\begin{table}[t!]
\centering

\begin{subtable}[t]{\linewidth}
\centering
\begin{tabular}{c c c l l}
\toprule
ADL & \makecell{$R$} & 
\makecell{Subtraction \\ Term} &
\makecell{GT Loc. (\%)} & 
\makecell{Top-1 \\ Loc. (\%)} \\ 
\midrule
\multirow{3}{*}{N} & N& N& 42.49& 31.38 \\
 & N& Y& 47.59 $\uparrow$& 35.01 $\uparrow$ \\ 
 & Y& N& 53.40 $\uparrow$& 40.19 $\uparrow$ \\ \midrule
\multirow{3}{*}{Y} & N& N& 71.59& 53.01 \\
 & N& Y& 75.78 $\uparrow$& 54.28 $\uparrow$ \\ 
 & Y& N& 73.56 $\uparrow$& 53.94 $\uparrow$ \\
\bottomrule
\end{tabular}
\caption{Localisation results on CUB-200-2011 with VGG-GAP.}
\label{tbl:ablation}
\end{subtable}

\vspace{1em}

\begin{subtable}[t]{\linewidth}
\centering
\begin{tabular}{c c c l l}
\toprule
ADL & \makecell{$R$} & 
\makecell{Subtraction \\ Term} &
\makecell{GT Loc. (\%)} & 
\makecell{Top-1 \\ Loc. (\%)} \\ 
\midrule
\multirow{3}{*}{N} & N& N& 54.56& 40.55 \\
 & N& Y& 54.29 $\downarrow$& 40.51 $\downarrow$ \\ 
 & Y& N& 57.73 $\uparrow$& 43.34 $\uparrow$ \\ \midrule
\multirow{3}{*}{Y} & N& N& 52.66& 36.88 \\
 & N& Y& 52.52 $\downarrow$& 37.08 $\uparrow$ \\ 
 & Y& N& 54.15 $\uparrow$& 37.76 $\uparrow$ \\
\bottomrule
\end{tabular}
\caption{Localisation results on CUB-200-2011 with ResNet50.}
\end{subtable}

\caption{Ablation study results on the importance of the region parameter $R$ and the subtraction term within the formulation of infoCAM.  Y and N indicates the use of corresponding feature. Arrows indicates the relative performance against the case where both features are not used.}
\label{table:ablation_study}
\end{table}

\subsection{Localisation Examples from Tiny-ImageNet}
We present examples from the Tiny-ImageNet dataset in \autoref{fig:cam-illustration-butterfly}. Such examples show the infoCAM draws tighter bound toward target objects.

\begin{figure*}[t!]
    \centering
    \includegraphics[width=\linewidth]{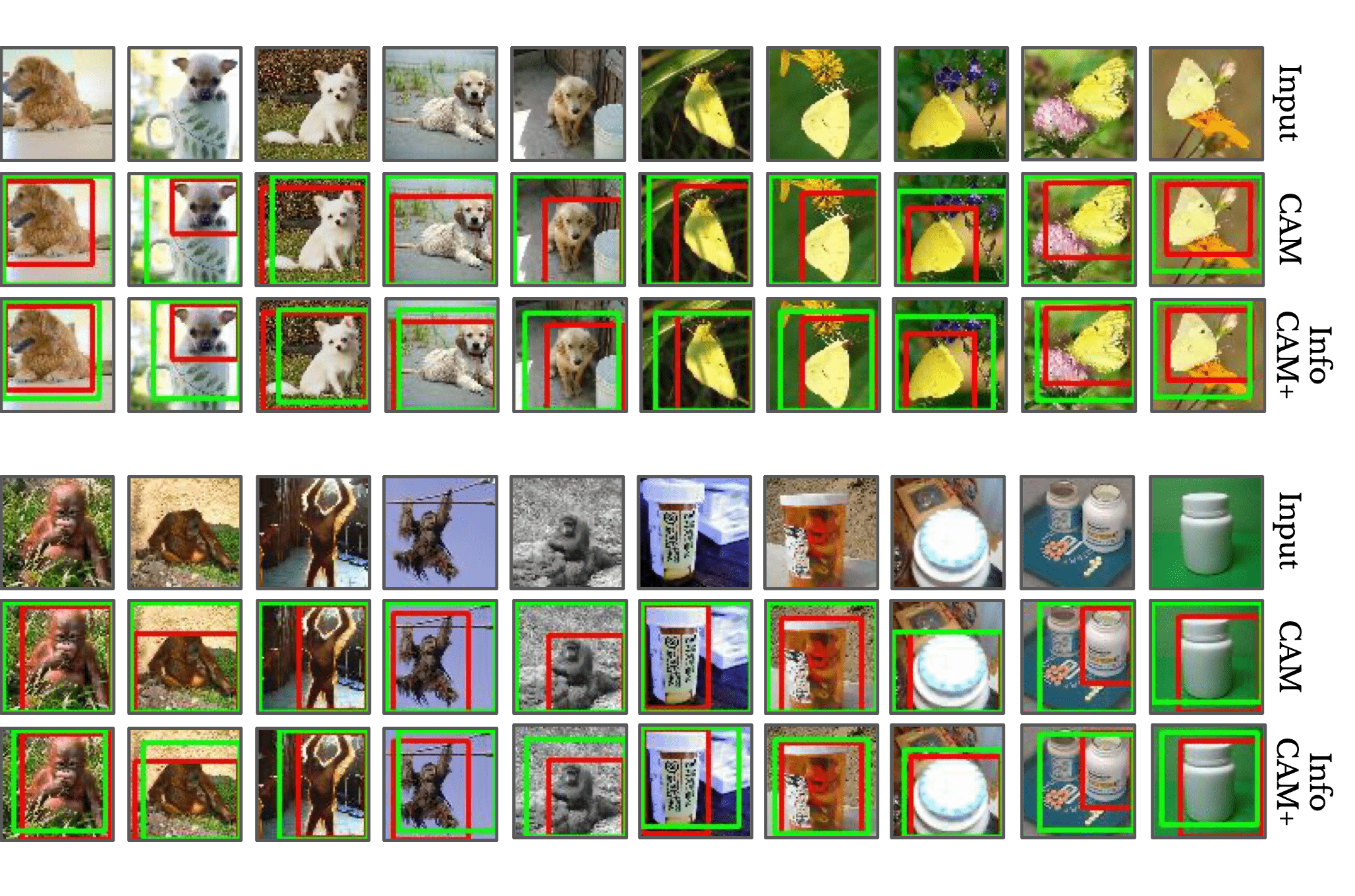}
    \caption{Visualisation of localisation with ResNet50 on CUB-200-2011 and TinyImageNet, without the assistance of ADL. The images in the second row are generated from the original CAM approach and the ones in the third row correspond to infoCAM. The red and green bounding boxes are ground truth and estimations, respectively. }
    \label{fig:cam-illustration-butterfly}
\end{figure*}

\end{document}